# Neuromorphic Digital-Twin-based Controller for Indoor Multi-UAV Systems Deployment

Reza Ahmadvand, Sarah Safura Sharif, Yaser Mike Banad*

School of Electrical and Computer Engineering, University of Oklahoma, 73019 USA
(*Corresponding author: bana@ou.edu).

*Abstract*—Presented study introduces a novel distributed cloud-edge framework for autonomous multi-UAV systems that combines the computational efficiency of neuromorphic computing with nature-inspired control strategies. The proposed architecture equips each UAV with an individual Spiking Neural Network (SNN) that learns to reproduce optimal control signals generated by a cloud-based controller, enabling robust operation even during communication interruptions. By integrating spike coding with nature-inspired control principles inspired by Tilapia fish territorial behavior, our system achieves sophisticated formation control and obstacle avoidance in complex urban environments. The distributed architecture leverages cloud computing for complex calculations while maintaining local autonomy through edge-based SNNs, significantly reducing energy consumption and computational overhead compared to traditional centralized approaches. Our framework addresses critical limitations of conventional methods, including the dependency on pre-modeled environments, computational intensity of traditional methods, and local minima issues in potential field approaches. Simulation results demonstrate the system's effectiveness across two different scenarios. First, the indoor deployment of a multi-UAV system made-up of 15 UAVs. Then the collision-free formation control of a moving UAV flock including 6 UAVs considering the obstacle avoidance. Owing to the sparsity of spiking patterns, and the event-based nature of SNNs in average for the whole group of UAVs, the framework achieves almost 90% reduction in computational burden compared to traditional von Neumann architectures implementing traditional artificial neural networks.

## I. INTRODUCTION

The evolution of autonomous robotic systems, particularly in multi-agent scenarios, faces significant challenges in managing task complexity, optimizing energy consumption, and ensuring safe operation in complex environments. While swarm intelligence and nature-inspired approaches have demonstrated remarkable potential in decentralized control [1], the implementation of these systems often struggles with computational efficiency and real-time processing demands, particularly when scaling to larger swarms. Simultaneously, advances in neuromorphic computing systems [2] present promising solutions to these computational challenges, suggesting a potential synergy between nature-inspired control architectures and efficient computing paradigms for distributed multi-agent systems. Traditional centralized control methods impose significant constraints on implementing sophisticated multi-agent systems, particularly in urban environments where complex obstacles and dynamic conditions prevail [3, 4]. While cloud-based control architectures [5, 6] have emerged as potential solutions, they introduce additional complexities, including the requirement for persistent plant-cloud communication and potential vulnerabilities in case of communication failures. These limitations become particularly critical in multi-UAV systems, where reliable operation must be maintained even in cases of central computer failure [3, 4]. The need for robust, decentralized control solutions that can maintain performance even with limited or interrupted cloud connectivity has become increasingly apparent. Spiking neural networks (SNNs) have emerged as a compelling solution for energy-efficient, low-latency control applications [7], while nature-inspired control methods have demonstrated exceptional capability in handling complex multi-agent scenarios [8, 9]. Traditional approaches to multi-UAV control face several critical limitations. Path planning and geometric guidance methods require a priori knowledge of the implementation environment with pre-modeled obstacles, limiting their effectiveness in dynamic or unknown environments [8, 9]. Potential field function approaches, while offering reactive control capabilities [10], can suffer from local minima problems and may not guarantee optimal trajectories. Model predictive control methods [11, 12], though capable of anticipating future states, introduce significant computational overhead due to their requirement for continuous prediction of system dynamics over future time intervals. Despite recent advances combining standalone path planning with particle swarm optimization (PSO) methods [13, 14], these hybrid approaches still struggle with real-time adaptation to dynamic obstacles [4]. Furthermore, while artificial potential field (APF) methods have shown promise in environments with known obstacle maps [15, 16, 17, 18], their effectiveness diminishes in scenarios with dynamic obstacles and multiple moving agents. These limitations become particularly apparent in urban environments where complex obstacles, including both static structures and dynamic obstructions, require real-time adaptation and decision-making capabilities. Inspired by digital twin (DT) approaches, our research introduces a novel framework that synthesizes these parallel developments by implementing a distributed control architecture where each UAV is equipped with its own SNN-based controller. This approach combines the energy efficiency, low latency and robustness of SNNs with nature-inspired control strategies. The system leverages the cloud in a neuromorphic DT architecture for complex calculations and control signal generation while maintaining local autonomy through individual SNNs that can learn and reproduce these control signals at the edge for each UAV separately. Building upon previous works in SNN-based frameworks for estimation and control of dynamical systems [19, 20] and recent advances in multi-UAV control systems incorporating artificial potential field concepts [15, 16, 17, 18], we propose a distributed cloud-edge architecture where: 1) Each UAV is equipped with an individual SNN that learns to replicate control signals generated by a sophisticated controller operating on the cloud in the neuromorphic DT architecture. 2) The cloud controller implements a nature-inspired collision-free formation control strategy based on tilapia fish territorial behavior and obstacle avoidance [21, 22]. 3) Each SNN operates independent from the implemented control algorithms in the cloud, ensuring continued operation of SNN-based controller at the edge even during discontinuous cloud-edge communication. 4) The distributed architecture maintains the benefits of swarm intelligence for situational awareness of the agents while incorporating efficient neuromorphic computing at the individual agent level.

The foundation of our approach integrates multiple innovative elements form nature-inspired formation control drawing from biological models [4], individual SNNs implementing leaky integrate-and-fire (LIF) neurons [23] and obstacle avoidance strategies [24, 25, 26], implemented through neuromorphic computing. Our methodology uniquely combines rapid learning capabilities at the edge with nature-inspired control principles in the cloud. This distributed architecture ensures that each agent can maintain effective control even during temporary losses of cloud connectivity, while still benefiting from the complex computational capabilities of cloud-based controllers when available. Further, to assess the validity of our approach it has been utilized in two different multi-UAV scenarios. First, the collision free optimal indoor deployment of a multi-UAV system consists of 15 UAVs, then it has been applied into the collision-free formation control of a moving flock of 6 UAVs in the presence of obstacles. Obtained results demonstrated that the proposed strategy has an acceptable performance in the implementation of our controller architecture.

The remainder of this paper is structured as follows: Section 2 presents the theoretical framework, including the integration of distributed SNNs with nature-inspired control strategies, multi-UAV dynamic modeling, and the underlying principles of both cloud-based and edge-based control. Section 3 details our implementation methodology, SNN architecture, and experimental setup. Section 4 presents comprehensive simulation results and analysis across various scenarios. Finally, Section 5 concludes with key findings and future research directions.

## II. THEORY

This section initiates with theoretical underpinnings and outcomes of our study, focusing on the application of SNNs in our proposed architecture and follows with controller design. Our approach leverages a nature-inpired controller on the cloud for collision-free optimal control of a multi-UAV system in a predefined barrier area [27]. While the controller on the cloud interacts with a virtual twin network (mathematical model on the cloud server which models the physical multi-UAV system) to produce the ideal output of the real-world multi-UAV system as an expected output for comparison with the obtained outputs of the navigation system in the monitoring center. In the presented approach each UAV has its own SNN-based controller that can learn to replicate the control signal $u$ provided by the cloud using a local online supervised learning approach [26, 28, 29]. Figure 1, depicts the proposed architecture for neuromorphic DT.

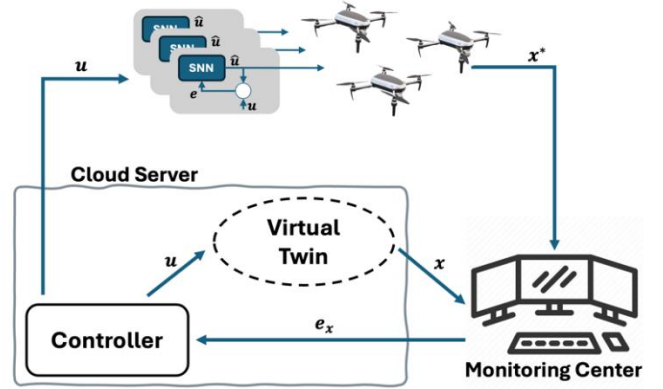

Fig. 1. Schematic of neuromorphic cloud-based DT

Moreover, our proposed architecture in this study leverages the advantages of neuromorphic computing such as energy efficiency of SNNs, compared to traditional computing frameworks such artificial neural networks (ANN) while it is using the optimality of the nature-inspired methods for the environmental awareness.

### A. SNN-based Edge Controller

Previously, a SNN-based cloud-edge framework for a single agent system has been introduced in [26]. The main concern of this section is to extend the previously proposed framework into a multi-UAV system. Using individual SNN-based controllers at the edge for each UAV. Thus, this section initiates with a brief review of the preliminaries. SNNs are the biologically inspired brain-mimicking computing tools that can efficiently replicate the neural circuits in the biological brain in which the neurons communicate via electrochemical signals called spikes. Based on the literature and engineering applications of SNNs, the focus of the presented study is on the utilizing networks consist of Leaky Integrate-and-Fire (LIF) neurons [30].

To develop a network of LIF neurons capable of replicating a desired signal like $u$, there are two strict assumptions that need to be satisfied. First, it is required to define a linear transformation like $u = Dr$, where $D$ is a fixed random decoding matrix that can extract the desired data from the filtered spike trains vector $r \in R^N$ [23, 30, 31, 32]. Filtered spike trains that obey the dynamic $\dot{r} = -\lambda r + s$ are the convolution of spike trains with synaptic kernels, and can be interpreted as instantaneous firing rate where $s \in R^N$ is emitted spike train by the neurons in each time step and has the faster dynamic than the $r$ [23]. Then the network works is a way to minize the cumulative prediction error which is the error between the actual value of $u$ and the estimated $\hat{u}$, by optimizing the timing of spike occurrence instead of changing the output kernel values $D$. To achieve this, it minimizes the following cost function [2, 5]:

$$J = \frac{1}{t}\int_0^t (\|\boldsymbol{u}(\tau) - \hat{\boldsymbol{u}}(\tau)\|_2^2 + \nu\|\boldsymbol{r}(\tau)\|_1 + \mu\|\boldsymbol{r}(\tau)\|_2^2 ) \, d\tau \quad (1)$$

Here, $\|.\|_2^2$ represents the Euclidean norm, and $\|.\|_1$ indicates L1 norms. The firing rule presented here, represents that the spike emmision is conditioned on the reduction of the predicted error, an analogy with predictive coding [31]. on the other hand, it is notable that, the neurons will emmit spike when their membrane potential touches a predefined threshold defined by $\boldsymbol{T}_i = (D_i^T D_i + \nu + \mu)/2$ that defines the thresholds for any individual neuron in the network where $D_i$ is the $i^{th}$ column of the matrix $D$, characterizing the neuron's output kernel and the change in the error due to a spike of $i^{th}$ neuron. And, The parameters $\nu$ and $\mu$ control the balance between efficiency and accuracy. Parameter $\nu$ pushes the network to be sparse (use minimal spikes), while $\mu$ promotes equal spike distribution across neurons. When these parameters are properly adjusted, the result is neural activity that's evenly spread throughout the network. Considering the ultimate goal of reducing the prediction error, the firing rule presneted here, minimizes the cost function $J$ by controlling the spike timing in response to level of excitation and inhibition [23, 31, 26]. As it has been introduced in [26], for the communication between UAVs and cloud, any individual UAV is required to be equipped with a SNN-based controller utilizing a recurrent network of LIF neurons defined by the following expression:

$$\dot{\boldsymbol{\sigma}} = -\lambda\boldsymbol{\sigma} - \Omega_f \boldsymbol{s} + \Omega_s \psi(\boldsymbol{r}) + kD^T \boldsymbol{e} \quad (2)$$

where

$$\dot{\Omega}_s = \eta\psi(\boldsymbol{r})(D^T \boldsymbol{e})^T \quad (3)$$

In the above expressions, $\boldsymbol{\sigma}$ refers to membrane potential, $\lambda$ represents the leak rate for the membrane potential, and $\boldsymbol{e} = \boldsymbol{u} - \hat{\boldsymbol{u}}$ is the network error for replicating the disered signal. $\psi(\boldsymbol{r}) = tanh(M^T \boldsymbol{r} + \boldsymbol{\theta})$ which is a highly unstructured nonlinear dendrite that can receive stochastic inputs from other neurons. where $M$ is the random fixed matrix, and $\boldsymbol{\theta}$ is a random vector that determines where the $tanh(.)$ changes its sign in a dendritic branch. Moreover, the above expression shows that the network utilizes two types of slow and fast synaptic connections. The slow connections $\Omega_s$ and fast connections $\Omega_f$. Slow connections control the main system behavior and dynamics, while the fast connections ensure spikes are distributed evenly throughout the network [26]. finally, using the network design introduced here the SNN-based controller at theedge can replicate the control signal provided by the cloud server. Figure 2 demonstrates the cloud-edge communication framework and the process of network weight updates for the UAVs during the mission. Using a predefined threshold for prediction error, each UAV autonomously updates its weight matrix, independent of other UAVs, to accurately reproduce its control input signal [26].

In the introduced strategy in [26], fro each UAV, the SNN-based edge controller undergoes two distinct phases of cloud communication. During the initial phase, spanning the first 50 time steps, the controller maintains continuous communication with the cloud. After this learning period, the system transitions to a more efficient operation mode where cloud communication is significantly reduced. To maintain performance and prevent significant deviations from the desired control signal, the edge controller receives updates from the cloud every 5 time steps. During these periodic updates, the controller obtains the reference signal, updates its internal state, and performs threshold-based weight adjustments using the latest error calculations. This optimized communication scheme results in a huge reduction in energy consumption for cloud-edge data transfer compared to the initial phase, making it highly efficient for practical implementations.

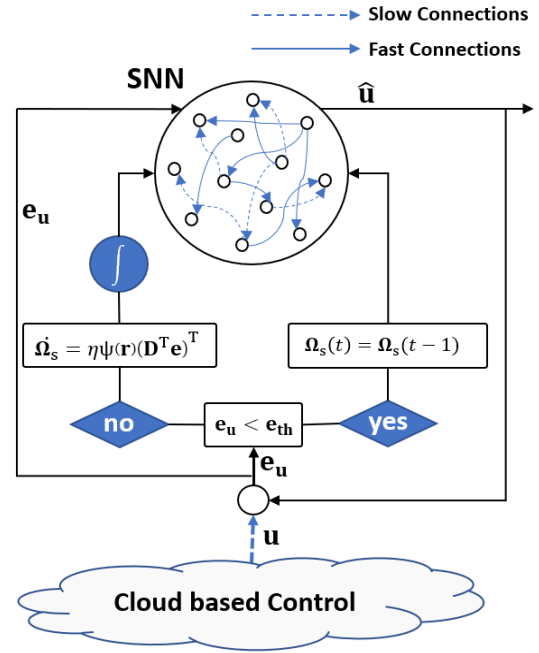

Fig. 2. Schematic diagram of weight update rule for SNN-based controllers at the edge [26].

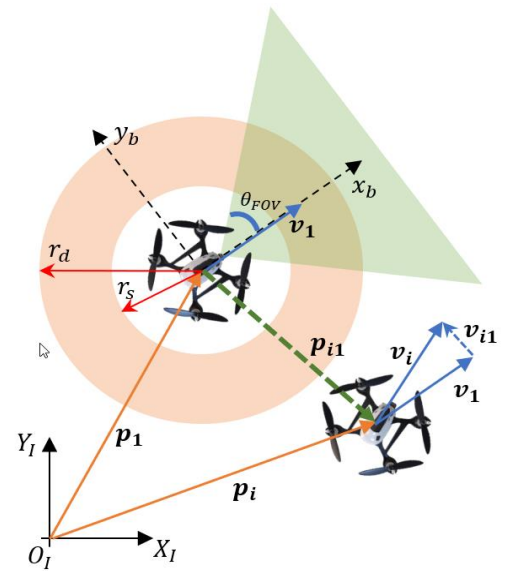

Fig. 3. Schematic design of multi-UAV system [27]

## B. Nature-Inspired Controller Design

This section focuses on the adopting the nature-inspired collision-free control approach proposed in [27] to the indoor deployment of a multi-UAV system running on our proposed neuromorphic DT-based architecture. Fig. 3 is demonstrating shcematic design of a multi-UAV system. In the presented design, to model the inertial translational motion of UAVs and their local dynamics, intertial coordinate frame, and body coordinate frames represented by $O_I X_I Y_I$ and $O_b x_b y_b$ have been leveraged respectively.

Considering the defined parameters in Fig. 3 and introducing $u_i$ as to the auxiliary control input for $i^{th}$ UAV, the following dynamical equations [27]:

$$\dot{p}_i = v_i \quad (4)$$
$$\dot{v}_i = u_i \quad (5)$$

where, $p_i$ and $v_i$ represent the position and velocity vectors of $i^{th}$ UAV. Paremeters, $p_{ij} = p_j - p_i$ and $v_{ij} = v_j - v_i$ are to relative position and velocity vectors. $r_s$ and $r_d$ are safety range (none of the UAVs are allowed to get closer to other UAVs than this range) and detection range respectively. $\theta_{FOV}$ is the field of view of the UAV with respect to the flight direction. Also, as it is common, to mathematically model a multi-UAV system for the purposes such as formation controller design and obstacle avoidance maneuvers, the UAVs in this study have been considered as particles [4, 33, 27].

To design the desired formation control law for the optimel deployment of a multi-UAV system consist of $N$ vehicles in a bounded area of space, the graph theory has been utilized. Thus, the graph $G(U, E)$ models a homogeneous system with a set of nodes defined as $U = \{i\}, i = 1,2,...,n$ and a set of $E = \{(i,j) | i,j \in U \ ; \ i \neq j\}$ that represents the edges which models the communication between the UAVs. Moreover, the following set of nodes in the space describes the neighorhood of each UAV in its detection range.

$$N_i = \{j \in U \setminus \{i\} \ | \ ||p_i - p_j|| < r_d\} \quad (6)$$

Furthermore, as introduced in [27] [34, 35, 36], ispired by tilapia fish territorial behavior, for an optimal deployment of UAVs in bounded spaces such as $Q \subset R$, leveraging a locational optimization, it has been proven that a centroidal Voronoi tessellation (CVT) is required in which the UAVs need to obey the following control law [27]:

$$u_{f_i} = -K_{p_i}(p_i - c_{v_i}) \quad (7)$$

In the above expression, $K_{p_i}$ is positive definite gain matrix which can be designed by trial and error, and $c_{v_i}$ refers to the centroid of each section in a CVT. Thus, to find the desired centroids for UAVs in a predefined bounded area, the probablistic generalized Lloyd's method has been leveraged. The pseudo-code of impelemented algorithm has been provided in Table 1 [27].

**Table 1.** Pseudo-code of the implemented Lloyd's CVT algorithm

**Probabilistic Generalized Lloyd's Algorithm**

*Input*:
- Predefined bounded area $Q \subset R$
- Density function $\rho(x)$ defined on $Q$
- Positive integer $N$ (number of UAVs)
- Positive integer $S_{num}$ (number of sampling points per iteration)
- Constants α1, α2, β1, β2 such that α1 + α2 = 1, β1 + β2 = 1, α2 > 0, β2 > 0

*Steps:*

*1. Initialization:*
- Choose an initial set of $N$ points $\{x_i\}^N_{i=1}$ in $Q$.
- Set iteration counters $\{j_i\}^N_{i=1} = 1$.

*2. Sampling:*
- Randomly sample $S_{num}$ points $\{y_r\}^{S_{num}}_{r=1}$ in $Q$ using a *uniform distribution* $\rho(x)$ as the probability density function.

*3. Point Update:*
  For each i = 1, 2, ..., $N$:
  - Gather all sampled points $y_r$ closest to $x_i$ (forming the set $W_i$, i.e., the Voronoi region of $x_i$).
  - If $W_i$ is empty, do nothing.
  - Otherwise:
    - Compute the average $u_i$ of the points in $W_i$.
    - Update $x_i$:
       $x_i \leftarrow ((α_1 j_i + β_1) / (j_i + 1)) x_i + ((α_2 j_i + β_2) / (j_i + 1)) u_i$
    - Increment $j_i$:
       $j_i \leftarrow j_i + 1$

*4. Repeat or Terminate:*
- Form the new set of points $\{x_i\}^N_{i=1}$.
- If $Q$ is a hypersurface, project $x_i$ onto $Q$
- Check stopping criteria (e.g., convergence or tolerance).
- If criteria are not met, go back to Step 2.

*Output:*
- Final set of points $\{x_i\}^N_{i=1}$.

It is notable that for the presented study, in the presneted algorithm in Table 1, the considered probability density function $\rho(x)$ is a uniform distribution which will lead to a uniformly distributed configuration of UAVs. Further, for considering the safety factors in the multi-UAV systems deployment, to avoid the inter-vehicle collision, and obstacle avoidance, the following parts has been added to the control law [27]:

- *Collision avoidance controller*

Inspired by Hook's law in the spring-mass-damper (SMD) system and considering relative position and velocities between the vehicles, the following expression has been introduced for the collision avoidance control law [27]:

$$u_{c_{ij}} = -k_{c1} p_c + k_{c2} v_{ij} \quad (8)$$

where:

$$p_c = \frac{1}{(||p_{ij}|| - r_s)^2} \frac{p_{ij}}{||p_{ij}||} \quad (9)$$

In the above expressions, $k_{c1}$ and $k_{c2}$ positive definite matrices.

- *Obstacle avoidance controller*

This part builds upon the obstacle avoidance controller design introduced in [27], which was inspired by how pigeons collectively maneuver around obstacles. We first review the original 2D planar model, where obstacles were limited to movement in the x direction. We then extend this approach to 3D space, which is the primary focus of our research. The method, consists of two main components: obstacle detection and maneuver execution. Fig. 4 is showing the schematic representation of obstacle detection.

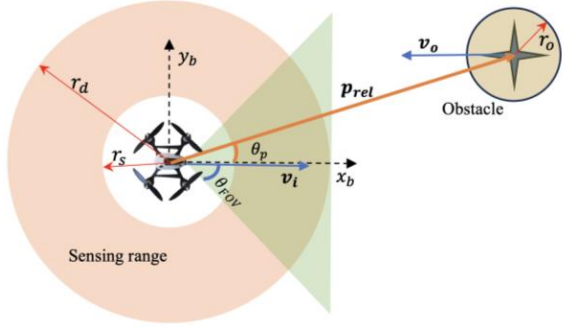

Fig. 4. Obstacle detection schematic representation [26]

Considerig the sector-like field of view (FOV) for the vehicles, and circular senesing range the two considtions have to be met. First, the osbtacle have to be in the sensing range, second it hjas to be in the FOV of the vehicles. Then the obstacle avoidanc needs to be performed. To this aim, the algorithm presented in Table 2 needs to be utilized.

Presented algorithm in Table 2, shows that in this study, for the obstacle avoidanc maneuver forbidden areas around the obstacles have been considered. Thus for the overal control law of the vehicles the following expression has to be implemented:

$$u_i = u_{fi} + u_{ci} \quad (10)$$

Please complete the sentence. Suggested edition: Through the above expression, each UAV obtains its control input vector in a semi-distributed fashion, where Lloyd's algorithm defines the desired spatial positions of the UAVs.

### III. NUMERICAL SIMULATION

This section presents the obtained results from the implementation of the neuromorphic DT framework into the indoor deployment, and a collision-free formation control scenario of a multi-UAV system considering obstacle avoidance.

#### A. Case-study1: Indoor deployment of multi-UAV system

This section presents the simulation results for deploying 15 UAVs in a rectangular area, approximating an indoor surveying scenario. Simulations ran for 10 seconds with a 0.01-second step, using the numerical values in Table 3. Throughout this work, the elements of matrix *D* are sampled from a zero-mean Gaussian distribution with covariance 1.

Figure 5 shows the UAVs' optimal configuration in the enclosed space, computed via Lloyd's method. Each Voronoi partition's center serves as a target position for the UAVs' controller.

**Table 2.** Pseudo-code of obstacle detection and avoidance

**Obstacle detection and avoidance algorithm**

*# Initialize obstacle detection metric*
$OD_{metric_{range}}$ = Array of zeros with size of number of obstacles $K$
# Loop over obstacles
FOR i = 1 TO $K$ DO:
  *# Calculate distance to the obstacle*
  $OD_{metric_{range}}$ = Euclidean norm of $(p_j - p_{Obs_i})$ for $i^{th}$ obstacle with respect to $j^{th}$ vehicle
  *# Calculate FOV*
  $OD_{metric_{FOV}}$ = Absolute difference of angles between object-obstacle and velocity direction
  *# Check if obstacle is detected*
  IF ($OD_{metric_{range}}$ < (obstacle radius + $r_d$)) AND ($OD_{metric_{FOV}} < \theta_{FOV}$ ):
    Set $OD_{metric}$[i] = 1
  ELSE:
    Set $OD_{metric}$[i] = 0
  ENDIF
  *# While any obstacle is detected*
  WHILE sum($OD_{metric}$) > 0 DO:
    *# Adjust position to avoid obstacle*
    IF $p_y > p_{Obs_{i_y}}$:
      $p_x$ = Reduce magnitude of $p_x$ by *scaler*
      $p_y$ = Increase magnitude of $p_y$ by *scaler*
    ELSE:
      $p_x$ = Reduce magnitude of $p_x$ by *scaler*
      $p_y$ = Reduce magnitude of $p_y$ by *scaler*
    ENDIF
    Update $p$ to new adjusted values ($p_x, p_y$ , and original $p_z$)
    *# Recalculate range and FOV metrics*
    Update $OD_{metric_{range}}$ and $OD_{metric_{FOV}}$ with new values
    *# Check if obstacle is still detected*
    IF ($OD_{metric_{range}}$ < (obstacle radius + $r_d$)) AND ($OD_{metric_{FOV}} < \theta_{FOV}$ ):
      Set $OD_{metric}$[i] = 1
    ELSE:
      Set $OD_{metric}$[i] = 0
    ENDIF
  ENDWHILE
ENDFOR

**Table 3.** Numerical values utilized in the simulations

| Parameter | Value |
|---|---|
| $r_d(m)$ | 3 |
| $r_s(m)$ | 1 |
| $K_p$ | $diag([1,1,1])$ |
| $K_v$ | $diag([2,2,3.5])$ |
| $k_{c1}$ | 10 |
| $k_{c2}$ | 1 |
| $\theta_{FOV}$ (deg) | 60 |
| $\mu$ | 0.0001 |
| $\lambda$ | 0.0001 |
| $\nu$ | 0.0001 |
| $k$ | 500 |
| $\eta$ | 0.01 |

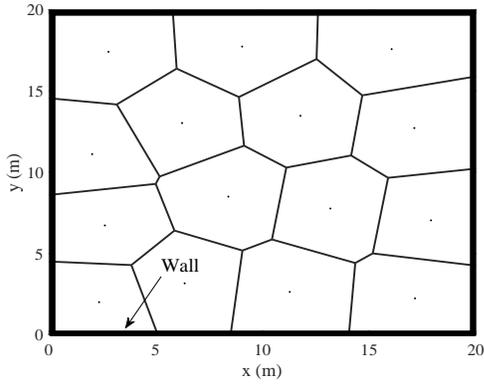

Fig. 5. Optimal formation of UAVs in the considered area obtained from Lloyd's method

Figure 6 shows the deployment results of a multi-UAV system using our proposed neuromorphic DT-based control architecture. It compares the predicted trajectories from a virtual twin—receiving control signals directly from the cloud-based controller—to those generated by an edge device model using an SNN-based controller. The close alignment of both trajectories indicates that the SNN-based approach at the edge successfully learns and applies the control signals expected from the virtual twin.

Figure 7 provides a more detailed comparison between the cloud-based controller's outputs and the SNN-based reproduction for UAV15. The SNN effectively replicates the desired control input, as evidenced by tracking errors converging to zero before $t=1$s with negligible fluctuations. This demonstrates the SNN's ability to learn and reproduce the control signals in an event-driven manner on edge devices.

Figure 8 illustrates the spiking pattern for UAV15. Most neural activity occurs before t=1 s (i.e., the first 100-time steps), corresponding to the interval during which the tracking errors in Figure 7 converge to zero. After that point, the network remains largely idle, reflecting the event-based nature of SNNs, which become active only when significant deviations exist between the reproduced and actual control signals. Quantitatively, the UAV15 controller uses 3,849 spikes—about 3.85% of the 100,000 available—confirming that SNNs reduce computational effort once tracking errors are minimal.

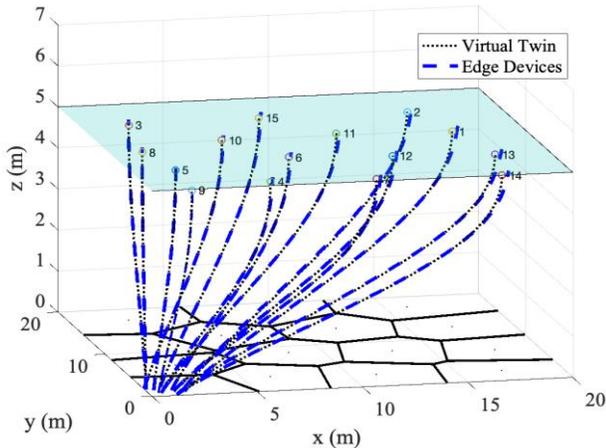

Fig. 6. Obtained trajectories for UAVs from virtual twins compared with obtained trajectories for edge devices utilizing signal reproduced by SNN-based controller. Video and Code are submitted in the supplementary.

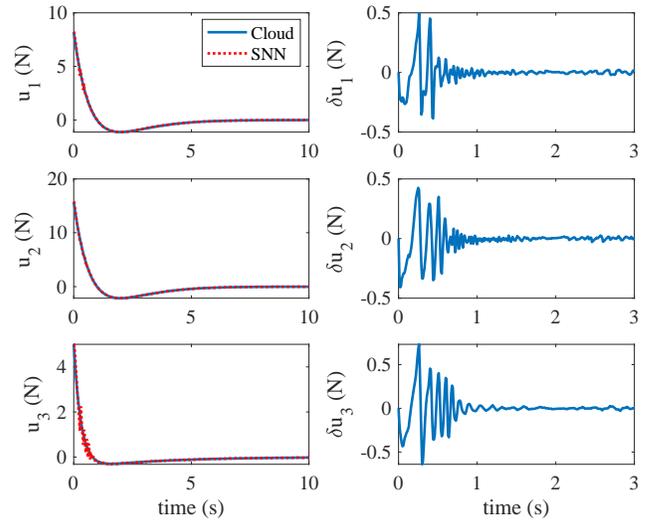

Fig. 7. Obtained trajectories for UAVs from virtual twin compared with obtained trajectories for edge devices utilizing signal reproduced by SNN-based controller

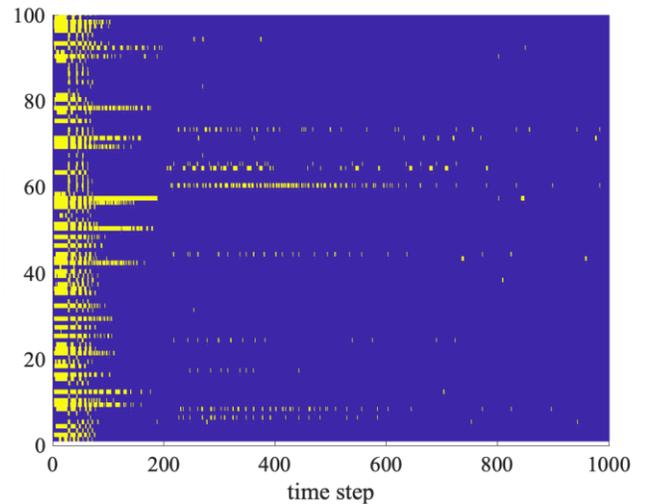

Fig. 8. Obtained trajectories for UAVs from virtual twin compared with obtained trajectories for edge devices utilizing signal reproduced by SNN-based controller

Table 4. Pseudo-code of obstacle detection and avoidance

| UAV | Number of Spikes | Resource Utilization (%) |
|---|---|---|
| 1 | 6110 | 6.1 |
| 2 | 7745 | 7.7 |
| 3 | 3060 | 3.1 |
| 4 | 1649 | 1.7 |
| 5 | 1700 | 1.7 |
| 6 | 2257 | 2.3 |
| 7 | 2266 | 2.3 |
| 8 | 2154 | 2.2 |
| 9 | 1328 | 1.3 |
| 10 | 2572 | 2.6 |
| 11 | 3258 | 3.3 |
| 12 | 2670 | 2.7 |
| 13 | 4801 | 4.8 |
| 14 | 3117 | 3.1 |
| 15 | 3849 | 3.8 |

It is worth noting that, because the results for all UAVs are qualitatively similar, only the data for UAV15 (chosen randomly) are presented here. To further analyze the proposed method's computational efficiency in a large-scale system of 15 UAVs, Table 3 provides the corresponding quantitative results. This table shows that the resource consumption for all UAVs ranges between approximately 1% and 8%, confirming the efficiency of our approach. In this architecture, the SNN-based controllers operate entirely independently of the cloud-based algorithm, eliminating the need to train separate flight computers on every controller parameter. Consequently, implementing our proposed architecture requires only minimal resources at the edge devices—an advantage in distributed multi-UAV operations.

### B. Case-study2: Moving flock of UAVs with obstacles

This section presents the obtained results from the implementation of our proposed architecture in the collision-free control of a moving flock of UAVs consist of 6 UAVs considering obstacles detection and avoidance maneuver. Simulation here has been conducted using the numerical values provided in Table 5 for 50 seconds with a time step of 0.01 second.

**Table 5.** Numerical values utilized in the simulations

| Parameter | Value |
|---|---|
| $r_d(m)$ | 1.5 |
| $r_s(m)$ | 0.5 |
| $K_p$ | $diag([1,1,1])$ |
| $K_v$ | $diag([2,2,2])$ |
| $k_{c1}$ | 3 |
| $k_{c2}$ | 1 |
| $\theta_{FOV}$ (deg) | 60 |
| $\mu$ | 0.001 |
| $\lambda$ | 0.005 |
| $\nu$ | 0.001 |
| $k$ | 500 |
| $\eta$ | 0.001 |

In the considered scenario here, firstly, the UAVs will form an optimal configuration obtained from the Lloyd's method in an area defined by $\{(x,y,z)|\ 0 < x < 2; 0 < y < 4; z = 5\}$, then they will move in the x-direction while they encounter to two static obstacles. Fig. 9 is illustrating the obtained optimal configuration for the UAVs flock in the desired area. Each point in the centroid of the Voronoi partitions has been assigned to any individual UAV.

Figure 10 compares the trajectories produced by the virtual twin model to those generated by the edge device models, which rely on SNN-based control signals. Overall, the plots illustrate how the edge devices successfully track the expected paths dictated by the virtual twin. Specifically, Fig. 10(a) shows the first phase of the scenario in which the UAVs converge to their optimal configuration (from Lloyd's method) with no inter-vehicle collisions. Fig. 10(b) captures the second phase, where the vehicles detect and avoid two separate obstacles, yet continue following their expected trajectories. In Fig. 10(c), the UAV formation recovers its optimal configuration without collisions, confirming the viability of the proposed architecture for multi-UAV surveying or mapping tasks. It is worth noting that, under closer scrutiny, the edge device trajectories exhibit minor deviations from those of the virtual twin. This indicates a small loss of accuracy in exchange for the computational efficiency of the SNN-based approach—an acceptable trade-off given the size of the surveyed area.

Figure 11 shows the spiking patterns for all UAVs in the moving flock. Although most networks display similar activity, UAVs 2 and 4—each encountering an obstacle in the x-direction—exhibit distinctly different behavior. After time step 2000, their neural firing increases sharply for a brief interval before returning to near-idle levels, whereas other UAVs remain largely idle throughout. This demonstrates the event-based nature of the SNN-based controllers, which adjust computational load and resource usage according to significant changes in input data. In contrast, a traditional approach such as an artificial neural network (ANN) would remain fully active at every time step, continuously consuming 100% of available resources. Moreover, in our architecture, each SNN operates independently, so elevated neural activity from UAVs 2 and 4 does not affect the other networks.

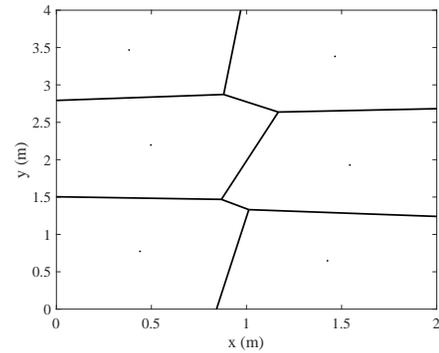

Fig. 9. Optimal configuration for the UAVs from the Lloyd's method.

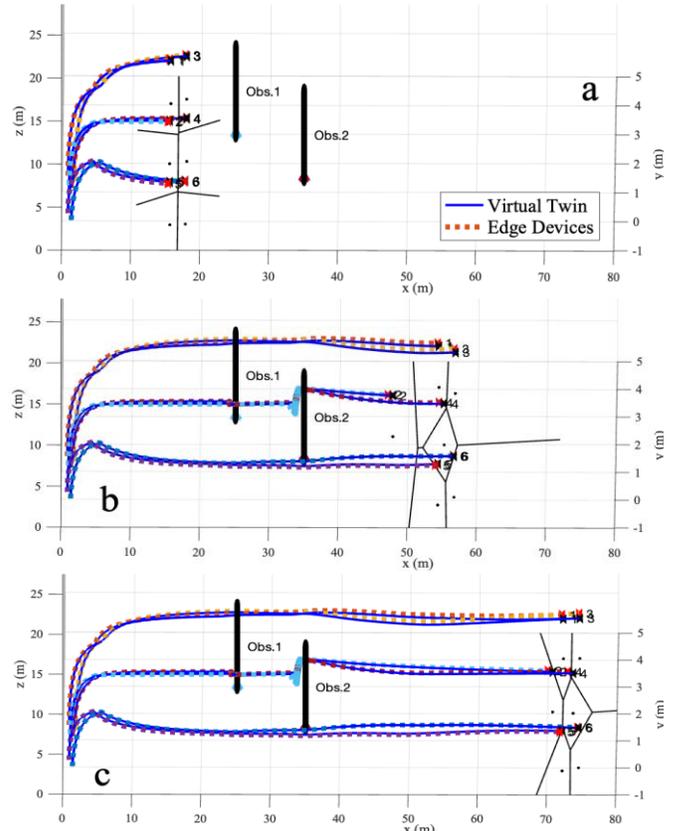

Fig. 10. Obtained trajectories for UAVs from virtual twin compared with obtained trajectories for edge devices utilizing signal reproduced by SNN-based controller for the moving UAV flock.

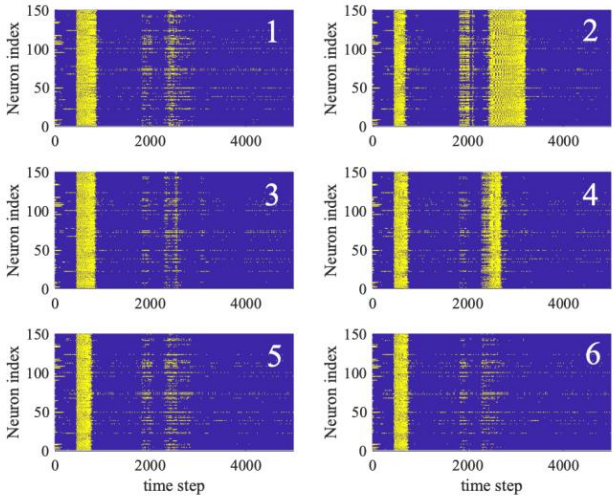

Fig. 11. Obtained spiking patterns for the UAVs in the moving flock.

**Table 6.** Pseudo-code of obstacle detection and avoidance

| UAV | Number of Spikes | Resource Utilization (%) |
|---|---|---|
| 1 | 33730 | 4.5 |
| 2 | 73583 | 9.8 |
| 3 | 31362 | 4.2 |
| 4 | 39015 | 5.2 |
| 5 | 26067 | 3.5 |
| 6 | 23304 | 3.1 |

To qualitatively assess the resource consumption of UAVs in this scenario, Table 6 lists the number of spikes used by each SNN-based controller to reproduce the desired control signals. Given the SNN size and total simulation time, up to 750,000 spikes are available. As shown in Table 6, UAVs that did not encounter obstacles used between 3 % and 4.5 % of their possible spikes, whereas UAV 2 and UAV 4—both of which interacted with obstacles—consumed around 10 %. These findings quantitatively confirm the spiking-pattern observations, where additional neural activity arises in response to obstacles, leading to higher resource usage.

## IV. CONCLUSION

The proposed distributed cloud-edge architecture for multi-UAV systems offers a transformative approach to addressing the challenges of task complexity, energy efficiency, and real-time adaptation in dynamic and obstacle-rich environments. By integrating spiking neural networks (SNNs) with nature-inspired control strategies, the framework achieves a robust balance between local autonomy and the computational capabilities of cloud-based controllers. The architecture's ability to maintain effective operation during temporary cloud connectivity losses demonstrates its resilience and adaptability, while its use of neuromorphic computing ensures energy efficiency and low-latency performance.

Experimental results in two distinct scenarios—collision-free indoor deployment of 15 UAVs and formation control of a moving flock of 6 UAVs—highlight the framework's effectiveness in achieving optimal, collision-free operations in complex environments. The synergy between SNN-based local controllers and the nature-inspired, cloud-driven global strategies enables scalable, decentralized control that can adapt to dynamic conditions in real time. This approach represents a significant step forward in the design of multi-agent autonomous systems, particularly for applications in urban and other complex operational domains. Future work may focus on further optimizing the framework's scalability and exploring its potential in other multi-agent contexts.